\documentclass[pdflatex,sn-mathphys-num]{sn-jnl}

\usepackage{graphicx}%
\usepackage{subcaption}
\usepackage{multirow}%
\usepackage{amsmath,amssymb,amsfonts}%
\usepackage{amsthm}%
\usepackage{mathrsfs}%
\usepackage[title]{appendix}%
\usepackage{xcolor}%
\usepackage{textcomp}%
\usepackage{manyfoot}%
\usepackage{booktabs}%
\usepackage{algorithm}%
\usepackage{algorithmicx}%
\usepackage{algpseudocode}%
\usepackage{listings}%
\usepackage{tabularx}

\usepackage{hyperref}
\usepackage{dsfont}
\usepackage{xspace}
\usepackage{rotating}
\usepackage[table]{xcolor}
\usepackage{pdflscape}
\usepackage{adjustbox}
\usepackage{colortbl}

\usepackage{epsfig}
\usepackage{lineno}%

\theoremstyle{thmstyleone}%
\theoremstyle{thmstyletwo}%

\theoremstyle{thmstylethree}%

\newcommand{\onedot}{.\xspace}
\newcommand{\myL}[1]{\ensuremath{\mathbf{L}_{#1}}}
\newcommand{\mycha}{\textbf{SPP}\xspace}
\newcommand{\mychb}{\textbf{SPI}\xspace}

\definecolor{GreenCell}{HTML}{D4F4DD}
\definecolor{RedCell}{HTML}{F8D2D2}

\newcommand{\yes}{\cellcolor{gray!10}\checkmark}
\newcommand{\no}{\cellcolor{white}$\times$}

\def\httilde{\mbox{\tt\raisebox{-.5ex}{\symbol{126}}}}
\def\sota{state-of-the-art }
\def\up{↑}
\def\down{↓}

\newcommand{\xTwoD}[4]{\mathbf{x}_{#1,#2,#3}^{#4}}         
\newcommand{\xTwoDhat}[4]{\hat{\mathbf{x}}_{#1,#2,#3}^{#4}} 
\newcommand{\xThreeD}[3]{\mathbf{X}_{#1,#2}^{#3}}          
\newcommand{\xThreeDhat}[3]{\hat{\mathbf{X}}_{#1,#2}^{#3}}

\def\etal{\emph{et al}\onedot}

\begin{document}

\title[Article Title]{A Multi-View Pipeline and Benchmark Dataset for 3D Hand Pose Estimation in Surgery}

\author*[1,2]{Valery Fischer}\email{vfischer@ethz.ch}
\author[1,2]{Alan Magdaleno}
\author[1]{Anna-Katharina Calek}
\author[1]{Nicola Cavalcanti}
\author[1]{Nathan Hoffman}
\author[1]{Christoph Germann}
\author[1]{Joschua Wüthrich}
\author[1]{Max Krähenmann}
\author[1]{Mazda Farshad}
\author[1]{Philipp Fürnstahl}
\author[1]{Lilian Calvet}

\affil[1]{\orgname{University Hospital Balgrist, University of Zurich}, \orgaddress{\country{Switzerland}}}
\affil[2]{ETH Zürich, Switzerland}

\abstract{
\textbf{Purpose} Accurate 3D hand pose estimation supports surgical applications such as skill assessment, robot-assisted interventions, and geometry-aware workflow analysis. However, surgical environments pose severe challenges, including intense and localized lighting, frequent occlusions by instruments or staff, and uniform hand appearance due to gloves, combined with a scarcity of annotated datasets for reliable model training.

\textbf{Method} We propose a robust multi-view pipeline for 3D hand pose estimation in surgical contexts that requires no domain-specific fine-tuning and relies solely on off-the-shelf pretrained models. The pipeline integrates reliable person detection, whole-body pose estimation, and state-of-the-art 2D hand keypoint prediction on tracked hand crops, followed by a constrained 3D optimization. In addition, we introduce a novel surgical benchmark dataset comprising over 68,000 frames and 3,000 manually annotated 2D hand poses with triangulated 3D ground truth, recorded in a replica operating room under varying levels of scene complexity.

\textbf{Results} Quantitative experiments demonstrate that our method consistently outperforms baselines, achieving a 31\% reduction in 2D mean joint error and a 76\% reduction in 3D mean per-joint position error.

\textbf{Conclusion} Our work establishes a strong baseline for 3D hand pose estimation in surgery, providing both a training-free pipeline and a comprehensive annotated dataset to facilitate future research in surgical computer vision.
}





\maketitle


\section{Introduction}

\begin{figure}[t]
\centering
\includegraphics[width=\linewidth]{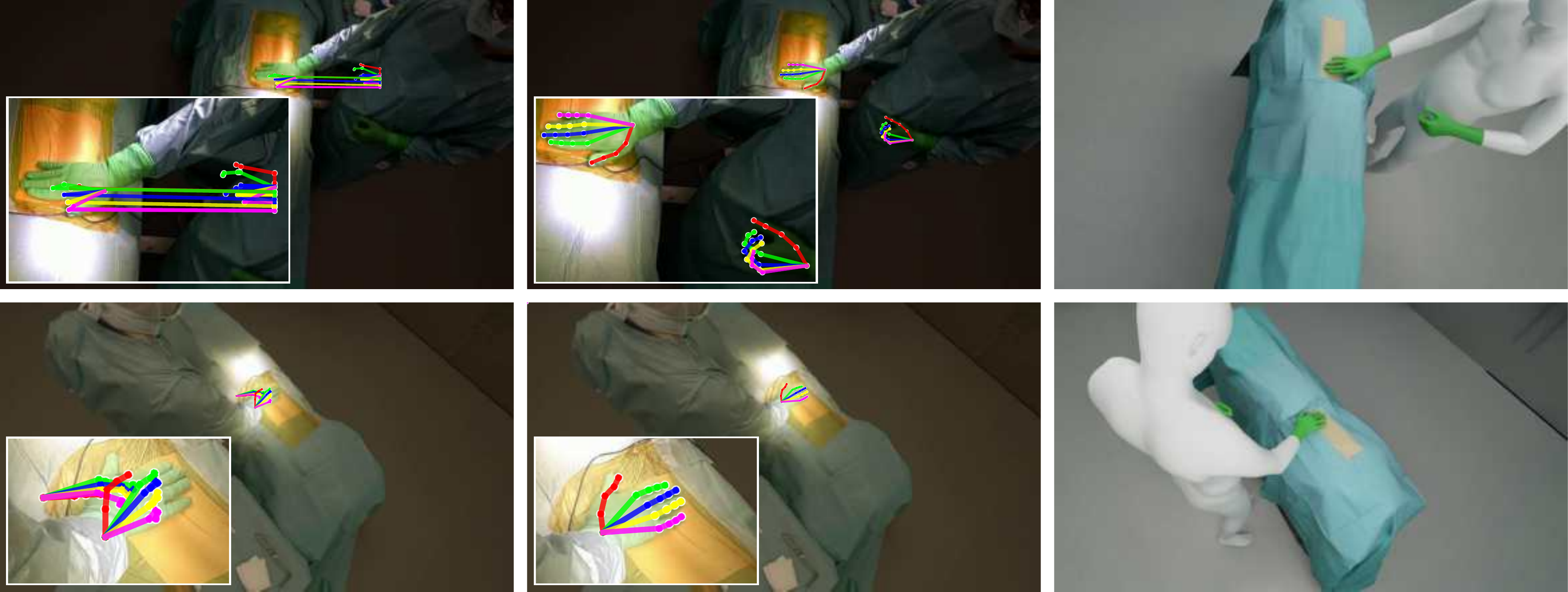}
\vspace{-3mm}
\caption{Example showing the challenges of hand localization in surgical scenes caused by lighting conditions and the similar appearance of gloves and gowns. Left: RTMPose~\cite{jiang2023rtmposerealtimemultipersonpose} fails to accurately localize the hands. Middle: our 2D predictions. Right: SMPL-X \cite{SMPL-X:2019} 3D reconstruction using our 2D predictions from the same viewpoints.}
\label{fig:example_imgs}
\end{figure}

Automatic 3D hand pose estimation in the surgical environment opens up new possibilities for tracking surgical activity, enabling detailed analysis of hand motions that can be used for workflow recognition \cite{padoy2019machine}, skill assessment \cite{zhang2021}, and surgical training \cite{bkheet2023, hein2021markerlesssurgical}.
At the same time, 3D hand tracking has the potential to enable natural, touch-free interaction with surgical robots, imaging systems, and augmented or virtual reality platforms, while helping to preserve the sterile field by reducing or eliminating physical contact with control panels or non-sterile surfaces \cite{muller2022}.

Approaches to 3D hand pose estimation can be broadly categorized into monocular and multi-view methods.
Monocular methods \cite{zimmermann2017learningestimate3dhand, iqbal2018handposeestimationlatent} are mostly learning-based pipelines that train deep neural networks on large annotated datasets to directly regress 3D keypoints or model parameters from RGB images.
While powerful in controlled environments, these methods often exhibit limited accuracy in unseen scenarios and struggle to generalize to out-of-distribution environments such as surgical scenes.
This limitation can be mitigated by incorporating additional information from pose priors \cite{mueller2017ganeratedhandsrealtime3d, xie2024msmanoenablinghandpose}.

In contrast, multi-view methods \cite{Hewitt_2024, Avola_2022} leverage multiple synchronized viewpoints, making the problem well-posed and resolving the scale ambiguity inherent in monocular estimation.
They also demonstrate improved robustness to occlusion, making them the preferred solution for surgical applications where both robustness and accuracy are paramount.
Typically, these methods consist of two main stages:
(i) detecting 2D joint locations (\textit{keypoints}) in each view, and
(ii) reconstructing the 3D configuration by optimizing the 2D keypoints under geometric and, optionally, anatomical constraints \cite{spurr2020weakly}, where 2D hand poses are generally extracted using data-driven methods \cite{khirodkar2024sapiensfoundationhumanvision, jiang2023rtmposerealtimemultipersonpose}.

Existing methods for 2D keypoint extraction face significant challenges under surgical conditions: surgical lighting is very bright, focused on a small region, and produces high-contrast glare, as illustrated in Fig.~\ref{fig:example_imgs}. Occlusions are frequent, caused by instruments or other staff, and gloves create uniform hand appearance with reduced texture cues. Poor performances of state-of-the-art 2D pose estimation methods \cite{khirodkar2024sapiensfoundationhumanvision, jiang2023rtmposerealtimemultipersonpose} can be mostly attributed to the lack of suitable training datasets.
On one hand, synthetic datasets such as RHD \cite{zimmermann2017learningestimate3dhand} provide dense annotations but fail to capture real-world visual noise and occlusions, as hands are typically rendered fully visible.
On the other hand, most real datasets such as FreiHAND \cite{Freihand2019} and HandCo \cite{HandCoZimmermann} lack 3D ground truth in realistic, cluttered conditions, or are limited to simple, highly controlled acquisition setups, while \cite{hein2021markerlesssurgical} is restricted to egocentric views only. 
Similar domain shift issues occur for whole-body pose estimation models that include hand predictions \cite{jiang2023rtmposerealtimemultipersonpose, khirodkar2024sapiensfoundationhumanvision, yang2023effectivewholebodyposeestimation}.
Srivastav \etal~\cite{SRIVASTAV2022102525} proposed adapting instance detection and pose estimation models to surgical settings but are restricted to body poses only.

\begin{figure*}[t]
    \centering
    \includegraphics[width=\linewidth]{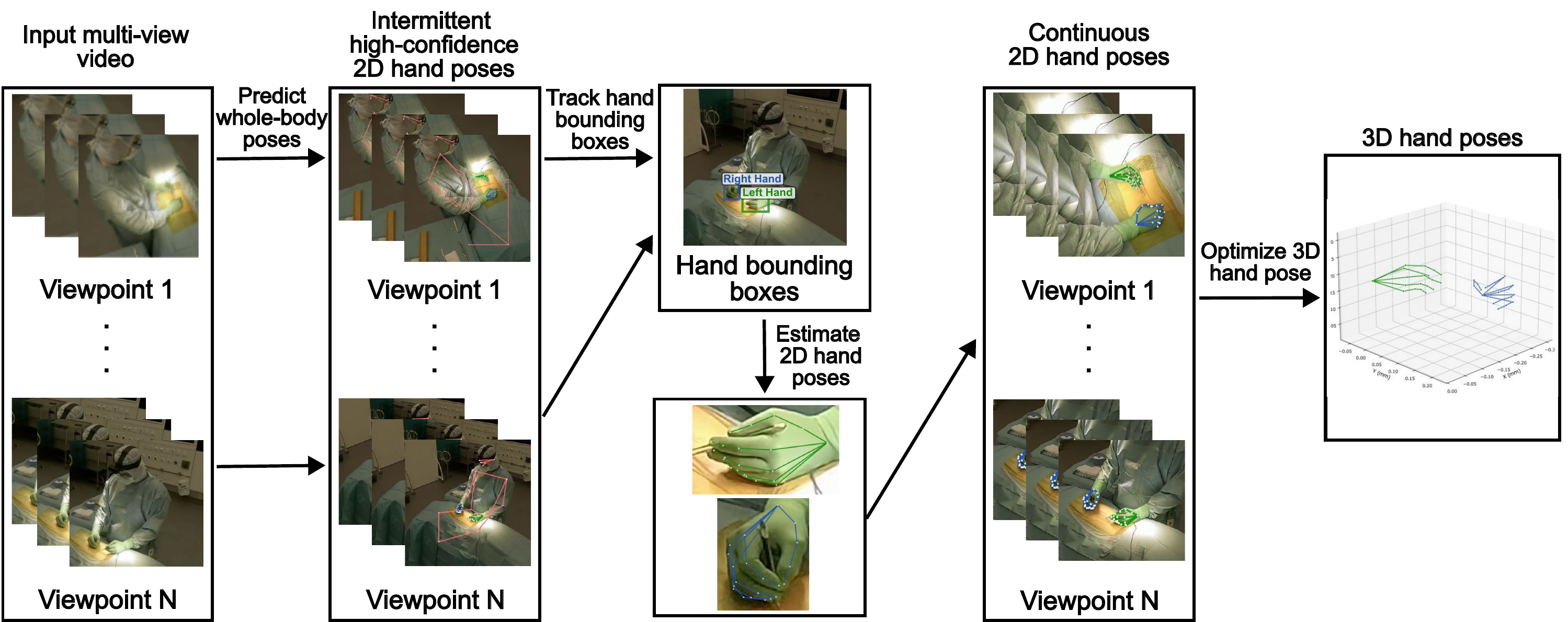}
    \caption{Overview of our 3D hand pose estimation framework. Given multi-view videos from the OR, we first detect and track persons, then predict full body and refined hand poses. Finally, we optimize a sequence of 3D hand poses consistent with anatomical and temporal constraints.}
    \label{fig:pipeline_overview}
\end{figure*}

In this work, we propose a robust, training-free pipeline for 3D hand pose estimation in surgical environments that overcomes domain shift and occlusion challenges without requiring retraining or fine-tuning.
Our design is motivated by three key observations:
(i) Modern object detectors such as YOLOv11 \cite{jocher_2025_15616841} can reliably detect humans even under unseen conditions.
(ii) Recent tracking models, notably Efficient Track Anything \cite{xiong2024efficienttrack}, enable robust long-term tracking under occlusions, appearance changes, and temporary detection failures. Their object-agnostic design makes them more resilient to domain shifts than hand-specific models, a property we explicitly leverage in surgical scenarios.
(iii) State-of-the-art 2D hand pose estimators achieve highly accurate joint localization when applied to well-cropped hand regions.

These insights motivate a top-down multi-stage pipeline, shown in Fig.~\ref{fig:pipeline_overview}, that extends the principle of top-down 2D body pose estimation, namely detecting persons first and then estimating poses inside their bounding boxes, to the robust detection and accurate localization of hand joints in surgical video without domain-specific retraining. The pipeline first detects individuals using YOLOv11, applies full-body 2D pose estimation within each bounding box, identifies keyframes with confident hand detections, and then tracks hands using Efficient Track Anything to maintain temporal consistency despite intermittent detections. Finally, hand keypoints are refined using a high-resolution 2D hand pose estimator.

From the resulting multi-view 2D detections, 3D hand poses are reconstructed through triangulation followed by constrained optimization. The optimization enforces reprojection consistency, temporal smoothness and shape consistency, producing robust and accurate hand trajectories.

In summary, our main contributions are threefold.
First, we present a modular, training-free pipeline for accurate multi-view 3D hand pose estimation in surgical environments.
Second, we introduce a novel real multi-view surgical dataset comprising over 68,000 frames and 3,000 annotated 2D hand poses with corresponding triangulated 3D ground truth.
Third, we conduct a comprehensive evaluation, including quantitative and qualitative comparisons with state-of-the-art methods as well as detailed ablation studies.

\section{Method}
\subsection{Problem formulation}
 
We consider the task of estimating accurate 3D hand poses in multi-view surgical videos captured by a set of calibrated and synchronized RGB cameras.  
Let the set of $V$ synchronized cameras be denoted as $\mathcal{C} = \{C_1, \dots, C_V\}$, each providing a video stream $\mathcal{V} = \{V_1, \dots, V_V\}$ of resolution $W \times H$.  
The sequence contains $N$ synchronized frames, with $M_t$ unique hands visible at time $t$, each represented by $J$ anatomical joints.

At each time step $t$, the 3D position of joint $j$ for hand $m$ in world coordinates is denoted by $\xThreeD{t}{j}{m} \in \mathbb{R}^3$, and its 2D projection in camera view $v$ by $\xTwoD{t}{j}{v}{m} \in \mathbb{R}^2$.  
The projection function $\Pi_v(\cdot)$ encodes the camera intrinsics and extrinsics, such that $\xTwoD{t}{j}{v}{m} = \Pi_v(\xThreeD{t}{j}{m})$.  
Predicted quantities are written as $\xThreeDhat{t}{j}{m}$ and $\xTwoDhat{t}{j}{v}{m}$ for 3D and 2D joint locations, respectively.
The problem can be decomposed into two subproblems.

The first is \textit{2D hand pose estimation}, where given the set of multi-view RGB frames $\{V_t^v\}_{v=1}^{V}$, the goal is to detect and localize 2D hand joints for all visible hands.  
Each view $v$ provides a set of detected hand bounding boxes $\mathcal{B}_{t,v}$, where $b_{t,v}^k = (x_{\min}, y_{\min}, x_{\max}, y_{\max})$ denotes the $k$-th detected region.  
Within each bounding box, a 2D hand pose estimator predicts a set of joint keypoints $\mathcal{P}_{t,v}^k = \{\left(\xTwoDhat{t}{j}{v}{k}, c_{t,v,j}^k\right)\}_{j=1}^{J}$, where $c_{t,v,j}^k \in [0,1]$ represent their confidence scores.

The second is \textit{3D hand pose estimation}, where given the associated 2D joint detections $\xTwoDhat{t}{j}{v}{m}$ and calibrated camera parameters, the goal is to estimate the corresponding 3D joint configuration $\mathcal{Q}_t^m = \{\xThreeDhat{t}{j}{m}\}_{j=1}^{J}$ for each hand $m$.

This formulation defines the 3D hand pose estimation task as the sequential resolution of 2D joint localization and 3D reconstruction under multi-view geometric consistency, which we adopt in this work.  
We describe our method for 2D pose estimation in Section~\ref{sec:2d_hand_pose}, and explore different sets of constraints from literature for the 3D pose estimation problem in Section~\ref{sec:opt}.

\subsection{2D hand pose estimation}
\label{sec:2d_hand_pose}

Figure \ref{fig:pipeline_overview} provides an overview of our pipeline.
In this section, we focus on the 2D pose estimation task (left part of the schematic), whose goal is to obtain accurate 2D hand keypoints in surgical scenes, where hands often appear small, are partially occluded, or undergo drastic illumination changes that obscure visual cues.
To address these challenges, we adopt a modular, top-down strategy that combines reliable body detection, robust tracking, and high-resolution hand pose refinement.
In the scope of this study, we focus on the case of a single surgeon in the scene.
The overall process comprises three main components:
(i) whole-body pose estimation for coarse initialization of hand locations,
(ii) hand tracking from confident detections using Efficient Track Anything to maintain continuous and robust hand bounding boxes, and
(iii) refined 2D hand pose estimation using a specialized model applied at near-native resolution for accurate localization.

\noindent \textbf{Whole-body pose estimation and initialization}
From synchronized multi-view input videos $\mathcal{V}$, we first detect the surgeon in each frame using YOLOv11~\cite{jocher_2025_15616841}, a state-of-the-art detector trained on large and diverse datasets.
Within each person bounding box, we apply a whole-body 2D pose estimator~\cite{jiang2023rtmposerealtimemultipersonpose} to predict full-body keypoints, including the hands.
These detections provide coarse hand locations and bounding boxes, which serve as initialization for the dedicated hand-tracking stage.
While these estimates may be imprecise due to the small image area of hands or strong lighting variations, they offer reliable priors for subsequent steps.

\noindent \textbf{Hand tracking using Efficient Track Anything}
\textbf{Keyframe selection.}
A set of keyframes in camera $v$ is chosen based on the average confidence $c_{t,v}^m$ of all hand joint confidence scores from the whole-body model.
For each hand, the frame with the highest $c_{t,v}^m > 0.3$ initializes tracking with Efficient Track Anything (TAM), which is then propagated forward and backward.
When tracking fails or reaches sequence boundaries, all successfully tracked frames for that hand are removed from the set, and the process restarts from the next highest-confidence frame until no valid detections remain.
This propagation maintains consistent hand identities and continuous localization even during occlusions, rapid movements, or temporary detection failures.
This process results in robust and temporally coherent hand trajectories $\mathcal{B}_{t,v}$ throughout the sequence.

\noindent \textbf{Accurate 2D hand pose estimation}
Each tracked hand region, cropped according to its propagated bounding box $\mathcal{B}_{t,v}$, is passed to a specialized 2D hand pose estimator for fine-grained keypoint prediction $\xTwoDhat{t}{j}{v}{m}$.
Operating under or near the native image resolution allows the model to recover detailed joint locations that are lost in the whole-body estimation stage.
This refinement step effectively mitigates the low-resolution limitation of whole-body methods and provides precise, temporally stable 2D hand keypoints that serve as high-quality input for the subsequent multi-view 3D hand pose optimization. In order to mitigate propagation of suboptimal 2D hand poses into 3D, individual keypoints are filtered by the confidence scores produced by the hand model. Joint predictions associated with confidence scores $c_{t,v,j}^k < 0.1$ are discarded.

\pagebreak


\subsection{3D hand pose estimation}
\label{sec:opt}

To obtain temporally consistent 3D hand trajectories, we formulate the 3D pose estimation problem as a constrained optimization problem solved with the L-BFGS-B algorithm. 
We incorporate constraints from prior work, some previously used in other contexts, including temporal smoothness, shape consistency to ensure stable hand geometry, and optionally biomechanical priors.
Given initial 3D hand poses derived from a canonical hand attached to the triangulated body pose, we refine them by minimizing a weighted sum of loss terms that encode these constraints:
\begin{equation}
L_{\text{total}} =
\lambda_{\text{reproj}} L_{\text{reproj}} +
\lambda_{\text{smooth}} L_{\text{smooth}} +
\lambda_{\text{shape}} L_{\text{shape}} +
\lambda_{\text{biomech}} L_{\text{biomech}}
\end{equation}
where each loss term is described below.
Each frame is optimized sequentially, using the result of the previous frame.

For all subsequent equations, the current frame is indexed by $i$.  
Let joint index $j=1$ correspond to the wrist.  
We define the centered 3D hand pose as
$\hat{\mathbf{X}}^{\text{centered}}_i = \hat{\mathbf{X}}_i - \hat{\mathbf{X}}_{i,1}$
that is, the 3D hand pose with the wrist joint translated to the origin.  
For hand $m$ and joint $j$, the corresponding centered keypoint is denoted as $\hat{\mathbf{X}}^{m,\text{centered}}_{i,j}$.

\noindent \textbf{Keypoint reprojection loss ($L_{\text{reproj}}$)}
This term enforces geometric consistency between the projected 3D joints and the observed 2D detections, similar to \cite{Avola_2022}, weighted by the confidence of the 2D predictions:
\begin{equation}
    L_{\text{reproj}} =
    \sum_{m=1}^{M} \sum_{v=1}^{V} \sum_{j=1}^{J}
    c_{i,v,j}^{m}
    \big\| \Pi_v(\xThreeDhat{i}{j}{m}) - \xTwoDhat{i}{j}{v}{m} \big\|_2^2.
\end{equation}

\noindent \textbf{Temporal smoothness loss ($L_{\text{smooth}}$)}
To encourage continuity over time, we penalize large changes in both joint positions and hand configuration between consecutive frames:
\begin{equation}
L_{\text{smooth}} =
\sum_{m=1}^M \sum_{j=1}^{J}
\big\| \xThreeDhat{i}{j}{m} - \xThreeDhat{i-1}{j}{m} \big\|_2^2.
\end{equation}

\noindent \textbf{Shape consistency loss ($L_{\text{shape}}$)}
This constraint ensures that the hand maintains a consistent shape across time by minimizing a rotation-invariant difference between consecutive frames, conceptually aligned with the method of Hewitt \etal~\cite{Hewitt_2024} for body pose:
\begin{equation}
L_{\text{shape}} = 
\big\| \hat{\mathbf{X}}^{\text{aligned}} - \hat{\mathbf{X}}^{\text{centered}}_{i-1} \big\|_F^2,
\text{ where }
\end{equation}
\begin{eqnarray*}
\hat{\mathbf{X}}^{\text{aligned}} = \hat{\mathbf{X}}^{\text{centered}}_i \mathsf{R}^T, \quad
\mathsf{R} = \mathsf{V}\mathsf{U}^T, \\
\mathsf{H} = \hat{\mathbf{X}}_i^{{\text{centered}},T}
\hat{\mathbf{X}}^{\text{centered}}_{i-1}, \quad
\mathsf{U}, \boldsymbol{\Sigma}, \mathsf{V}^T = \text{SVD}(\mathsf{H}).
\end{eqnarray*}
This corresponds to the orthogonal Procrustes problem, which computes the rotation 
$\mathsf{R}$ that best aligns the two centered point sets in the least-squares sense.

\noindent \textbf{Biomechanical loss ($L_{\text{biomech}}$)}
We also study the potential contribution of a biomechanical loss, although such terms are often found to be less effective. 
This loss encourages biomechanical plausibility through a differentiable function that enforces joint-angle limits and anatomical feasibility, following Spurr \etal~\cite{spurr2020weakly}.

\section{Benchmark dataset}

To evaluate our method, we introduce a novel dataset of surgical hand motions captured with a calibrated multi-view setup in a replica of an operating room that faithfully reproduces real surgical conditions.

The dataset comprises two recording sessions. The first, \mycha (spinal preparation), captures a single-surgeon procedure during the preparation phase using 12 synchronized static GoPro Hero Black 12 cameras (4K, 30~fps, fixed shutter). The second, \mychb (spinal instrumentation), records a multi-surgeon cadaveric spinal fusion surgery using 11 GoPro Hero Black 12 cameras. All cameras were synchronized through a custom PCB controller ensuring sub-frame alignment \cite{meyer2025rocsync}.
Internal and external calibrations were performed using the OpenCV library and~\cite{flückiger2025automaticcalibrationmulticameralimited}.

\noindent\textbf{Complexity levels}  
To systematically organize scene variability and support controlled benchmarking, we define a hierarchy of scene complexity levels, denoted as \myL{0} through \myL{5}. Each level progressively increases the difficulty of hand pose estimation by introducing factors such as occlusion, motion, and the number of participants.

\noindent\textbf{\myL{0} – Single-person (static):}  
A single person performs simple actions while remaining mostly stationary. All hand joints are fully visible in at least two views. This setting provides clean, unambiguous visual data ideal for baseline evaluation.

\noindent\textbf{\myL{1} – Dynamic single-person (slow, small tools):}  
A single surgeon performs movements with slow hand motion and uses small surgical tools. Moderate occlusions may occur, but the structure of the hands remains largely visible. This level introduces natural variability while preserving controlled conditions.

\noindent\textbf{\myL{2} – Dynamic single-person (fast, large tools):}  
Faster hand and arm motion combined with the use of larger tools causes stronger occlusions. These factors challenge temporal consistency and 3D reconstruction accuracy.

\noindent\textbf{\myL{3} – Static multi-person:}  
Several participants (up to three) remain within the field of view, with limited movement and minimal occlusions. The setting increases visual clutter but retains good observability across cameras.

\noindent\textbf{\myL{4} – Partial observability:}  
Multiple participants interact dynamically, often causing hands or tools to be partially hidden. Some camera views lose full hand visibility, introducing realistic visual ambiguity and limited multi-view coverage.

\noindent\textbf{\myL{5} – Dynamic multi-person:}  
The number of visible participants changes over time with frequent occlusions, entering and leaving the scene. This represents the most unconstrained and complex condition.

In this work, we focus on \myL{0}–\myL{2}, which capture increasing complexity. The higher complexity levels (\myL{3}–\myL{5}) are included for future benchmarking.

The dataset comprises 20 sequences (300 frames each) spanning complexity levels \myL{0}–\myL{5}, totaling over 68{,}000 synchronized multi-view frames.

\noindent\textbf{Annotation details}
For each sequence, we manually labeled 10 keyframes per camera, uniformly distributed across the 300-frame sequence, using a custom annotation tool. This process resulted in approximately 3{,}000 hand pose annotations. The 2D joint labels were triangulated into 3D ground truth using camera parameters, and their reprojections are used as the ground-truth 2D hand poses. Annotation quality was further verified and evaluated through multi-annotator cross-review to ensure consistency (see Section 2.4 of the supplementary material for details).

\section{Experiments}

We conduct a comprehensive evaluation of the proposed method on the 2D hand pose estimation task in isolation (using 2D and 3D metrics), and compare it against 2D hand pose baselines, evaluated either independently or in combination.
We also analyze the impact of the studied loss terms on 3D optimization through an extensive ablation study, all performed on the proposed dataset across complexity levels \myL{0} to \myL{2}.

\noindent \textbf{Baselines}  
Our 2D pose estimation evaluation includes both whole-body and hand-specific baselines. We consider Sapiens~\cite{khirodkar2024sapiensfoundationhumanvision}, RTMPose~\cite{jiang2023rtmposerealtimemultipersonpose} (noted RTM), and DWPose~\cite{yang2023effectivewholebodyposeestimation} (noted DW) as whole-body baselines. In addition, we evaluate a hand pose model from RTMPose~\cite{jiang2023rtmposerealtimemultipersonpose} (noted HM) —as used in our pipeline— applied to cropped images extracted from bounding boxes around the hand regions predicted by the whole-body models. This setup allows us to isolate and assess the contribution of the hand tracking component in our approach. The notation \mbox{$A \rightarrow B$} indicates that the whole-body method $A$ is used to obtain the hand bounding boxes, and the hand pose estimation model $B$ is then applied within those regions.

\noindent \textbf{Experimental scores} We evaluate our method using standard metrics that jointly assess 2D and 3D accuracy. The mean per joint position error (MPJPE) measures the average Euclidean distance between predicted and ground-truth 3D joint positions, while the mean reprojection error (MRE) quantifies how well the predicted 3D joints project back into the 2D image space. The percentage of correct keypoints in 2D ($PCK_{2D}@\tau$) and in 3D ($PCK_{3D}@\delta$) represent the fraction of keypoints whose reprojection or Euclidean error is below pixel or millimeter thresholds $\tau$ and $\delta$, respectively. We also report the mean PCK ($mPCK$), computed as the average of PCK scores across thresholds (5, 10, 20, 30 for 2D and 5, 10, 25, 50 for 3D), expressed in pixels or millimeters. For 2D detection performance, we compute the average precision (AP) following the COCO evaluation protocol~\cite{lin2015microsoftcococommonobjects}. Finally, the mean joint error (MJE) measures the average 2D Euclidean error between predicted and ground-truth joints. Together, these metrics provide a comprehensive evaluation of geometric accuracy, reprojection consistency, and keypoint localization in both 2D and 3D spaces.

\begin{table}[t]
\centering
\rowcolors{2}{gray!10}{white} 
\begin{tabular}{lccccccc}
\toprule
\textbf{Metric} & Sapiens & RTM & DW & Sapiens $\rightarrow$ HM & RTM $\rightarrow$ HM & DW $\rightarrow$ HM & Ours \\
\midrule
\textbf{MJE} $\downarrow$ & 29.5 & 27.5 & 23.5 & 20.9 & \underline{16.9} & 17.0 & \textbf{11.8} \\
$mPCK_{2D}$ $\uparrow$ & 7.7 & 12.6 & 14.0 & 17.4 & 20.8 & \underline{20.9} & \textbf{22.8} \\
\textbf{AP} $\uparrow$ & \underline{67.3} & 66.9 & 66.9 & 66.9 & 65.4 & 66.7 & \textbf{70.3} \\
\bottomrule
\end{tabular}
\caption{Quantitative evaluation on the 2D pose predictions, comparing different \sota methods against our method. HM refers to the RTM Hand Model used during pose refinement. \textbf{Best scores}, \underline{Second Best Scores}.}
\label{tab:2D_pose_results}
\vspace{-6mm}
\end{table}

\begin{figure}[t]
    ~~~~~~ \includegraphics[width=0.9\linewidth]{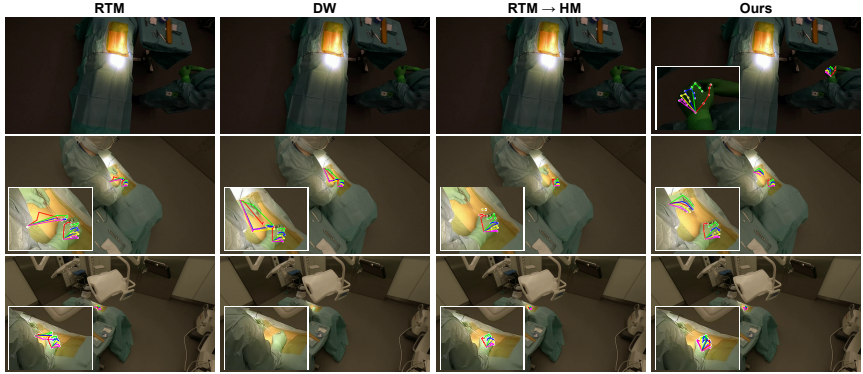}
    \caption{Qualitative comparison of 2D hand pose predictions. RTM, DW, and RTM $\rightarrow$ HM baselines are shown alongside our method, which achieves more complete and stable hand poses.}
    \label{fig:poses_2d_results}
\end{figure}

\noindent \textbf{2D pose results} Quantitative results are presented in Table~\ref{tab:2D_pose_results}, alongside our baseline comparisons. Figure~\ref{fig:poses_2d_results} illustrates representative qualitative examples comparing the baselines with our method. The results show that our method significantly outperforms state-of-the-art models in predicting 2D hand poses across all reported metrics. This demonstrates that our top-down approach, combined with the tracking component of the model (noted T), enables more accurate and stable 2D hand pose predictions. The improvement can largely be attributed to the tracker’s object-agnostic design, which allows it to localize hands effectively without prior knowledge of the target object. This leads to more reliable bounding box predictions and, consequently, improved 2D hand pose estimation. These results are further supported by the 3D evaluations presented in Table \ref{tab:hand_refinement_results_transposed}, where our method shows a clear advantage over predictions obtained from whole-body estimators alone (first row) or when combined with hand-specific models (second row).

\newcolumntype{Y}{>{\centering\arraybackslash}X}
\begin{table}[t]
\centering
\rowcolors{2}{gray!10}{white} 
\begin{tabularx}{\textwidth}{lYYYY}
\hiderowcolors
\toprule
\textbf{Method} & \textbf{MPJPE} $\downarrow$ &  \textbf{MRE} $\downarrow$ & $mPCK_{2D}$ $\uparrow$ & $mPCK_{3D}$ $\uparrow$ \\
\midrule
\showrowcolors
RTM~\cite{jiang2023rtmposerealtimemultipersonpose} & \underline{35.8} & \underline{39.8} & 39.8 & 40.8 \\
RTM $\rightarrow$ HM~\cite{jiang2023rtmposerealtimemultipersonpose} & 51.9 & 58.1 &  \underline{42.3} & \underline{41.4}\\
Ours: RTM + T~\cite{xiong2024efficienttrack} $\rightarrow$ HM~\cite{jiang2023rtmposerealtimemultipersonpose} & \textbf{8.5} & \textbf{8.2} & \textbf{76.6} & \textbf{74.7} \\
\bottomrule
\end{tabularx}
\caption{Quantitative results on the 2D pose predictions in 3D.
MPJPE (mm) and MRE (px): lower is better; $mPCK_{2D}$ and $mPCK_{3D}$: higher is better. 
Results correspond to the 3D optimization in Eq.~(1) with weighting factors 
$\lambda_{\text{reproj}} = 1.0$, $\lambda_{\text{smooth}} = 20.0$, 
$\lambda_{\text{shape}} = 50.0$, and $\lambda_{\text{biomech}} = 0$ (baseline configuration) for all 2D methods. 
\textbf{Best} and \underline{second best}.}
\vspace{-6mm}
\label{tab:hand_refinement_results_transposed}
\end{table}

\begin{table}[t]
\centering
\rowcolors{5}{gray!10}{white} 
\begin{tabularx}{\textwidth}{l | YYYY || YYYY}
\hiderowcolors
\toprule
Keypoint reprojection & \yes & \yes & \yes & \yes & \yes & \yes & \yes & \yes\\
Temporal smoothness & \no  & \yes & \yes & \yes & \no  & \yes & \yes & \yes \\
Shape consistency & \no & \no & \yes & \yes & \no & \no & \yes & \yes\\
Biomechanical & \no & \no & \no & \yes & \no & \no & \no & \yes\\
\midrule
\midrule
&  \multicolumn{4}{c||}{\textbf{MPJPE} $\downarrow$} & \multicolumn{4}{c}{$mPCK_{3D}$ $\uparrow$} \\
\midrule
\showrowcolors
Complexity level \textbf{L$_0$} & 115.1 & \textbf{8.0} & \textbf{8.0} & 24.0 & 59.5 & \textbf{75.7} & \underline{75.4} & 71.6\\
Complexity level \textbf{L$_1$} & 47.6 & \textbf{8.8} & \underline{9.1} & 14.4 & 64.5 & \textbf{74.8} & \underline{73.2} & 64.5\\
Complexity level \textbf{L$_2$} & 297.9 & \underline{9.5} & \textbf{8.8} & 69.7 & 55.9 & \underline{74.2} & \textbf{75.4} & 23.7  \\
Mean & 112.9 & \textbf{8.4} & \underline{8.5} & 25.9 & 60.4 & \textbf{75.1} & \underline{74.7} & 58.6 \\
\bottomrule
\end{tabularx}
\caption{Ablation study showing the effect of individual optimization terms on reconstruction accuracy. Active components are marked with \yes. The metrics reported are MPJPE and $mPCK_{3D}$ (mm), with L$_0$–L$_2$ denoting increasing scene complexity and the mean score over all scenes. \textbf{Best} and \underline{second best}.}
\vspace{-6mm}
\label{tab:abl_studies_errors}
\end{table}

\begin{figure*}[t]
    \centering
    \includegraphics[width=\linewidth]{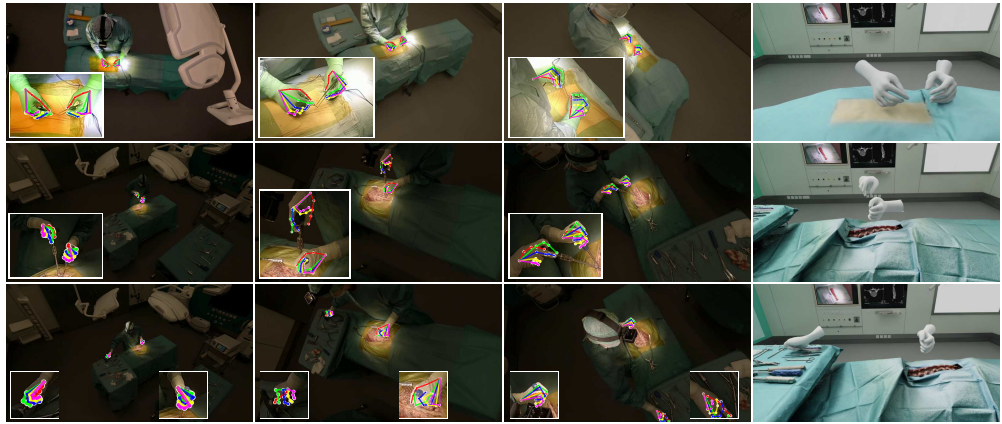} 
    \vspace{-4mm}
    \caption{Qualitative results of the full pipeline showing 3D poses reprojected into the original video streams and a 3D SMPL-X~\cite{SMPL-X:2019} render (right). First row: SPP. Second and third row: SPI.}
    \label{fig:qualitative_results}
\end{figure*}

\noindent \textbf{3D pose results}
Results are summarized in Table~\ref{tab:abl_studies_errors}, with qualitative examples shown in Figure~\ref{fig:qualitative_results}. Each variant of our optimization method is evaluated independently to analyze the impact of individual loss terms through an ablation study. 
As expected, the temporal smoothness constraint consistently yields significant improvements. 
The shape consistency term provides comparable performance on average but proves particularly effective for fast hand motions in $\mathbf{L}_2$ sequences, where it mitigates the degradation of 2D keypoints. 
Therefore, it is retained in our baseline configuration. 
In contrast, including the biomechanical constraint decreases performance, likely because the most plausible biomechanical shape does not precisely match the multi-view 2D predictions, leading to lower evaluation scores.

\section{Discussion}

\noindent \textbf{Limitations and future work}
Although the proposed multi-view pipeline demonstrates strong performance for 3D hand pose estimation in surgical environments, several challenges remain unaddressed. The current implementation focuses primarily on controlled single-person scenarios, while more complex settings involving multiple participants, frequent self-occlusions, and low hand visibility pose additional difficulties that our method does not yet handle effectively. Execution time is another limitation, as the optimization and tracking stages remain computationally expensive, requiring around 30–45 minutes for a 300-frame sequence on an RTX 2060 GPU. Future work will target faster, learning-based feedforward models trained using 3D annotations produced by our robust pipeline, paving the way to near real-time performance. Extending the system to multi-person and occlusion-heavy scenes also represent key research directions.

\noindent \textbf{Conclusion}
We presented a multi-view pipeline for 3D hand pose estimation in surgical settings that operates without domain-specific training. The integration of a top-down approach coupled with robust tracking and pose refinement yields superior accuracy compared to state-of-the-art methods, even under challenging occlusions. Despite current runtime limitations, this work establishes a solid foundation for future real-time, learning-based systems for surgical motion understanding.

\noindent \textbf{Acknowledgements} This work has been supported by the OR-X - a swiss national research infrastructure for translational surgery - and associated funding by the University of Zurich and University Hospital Balgrist.


\bibliography{sn-bibliography}

\def\httilde{\mbox{\tt\raisebox{-.5ex}{\symbol{126}}}}
\def\sota{state-of-the-art }
\def\up{↑}
\def\down{↓}

\clearpage

\appendix
\section*{Appendix}
\renewcommand{\thesection}{Appendix \Alph{section}}
\renewcommand{\thesubsection}{\Alph{section}.\arabic{subsection}}
~~~~~This supplementary material provides additional technical details,
extended experimental results, and comprehensive information about the proposed dataset that could not be included in the main manuscript due to space constraints. Section~\ref{sec:scores} defines the evaluation metrics used throughout our experiments. Section~\ref{sec:dataset} details the dataset and acquisition setup.
Additional qualitative and quantitative results are reported in Section~\ref{sec:additional_results}. Finally, Section~\ref{sec:impl_details} details implementation aspects and parameter settings for reproducibility.

\section{Evaluation scores}
\label{sec:scores}

To comprehensively assess the performance of our proposed method, we employ a set of standard quantitative metrics commonly used in hand and human pose estimation tasks.

\noindent\textbf{Mean Per Joint Position Error (MPJPE)}  
This metric measures the mean Euclidean distance between predicted and ground-truth 3D joint positions:
\begin{equation}
\text{MPJPE} = \frac{1}{NJ} \sum_{i,j,m=1}^{N,J}  \left\| \xThreeDhat{i}{j}{} - \xThreeD{i}{j}{} \right\|_2,
\end{equation}
where $N$ is the number of frames, $J$ is the number of joints, $\xThreeDhat{i}{j}{}$ is the predicted 3D position of joint $j$ in frame $i$, and $\xThreeD{i}{j}{}$ is the corresponding ground truth.
\\\\
\noindent\textbf{Mean Reprojection Error (MRE)}  
This metric evaluates how well the predicted 3D joints project back into the 2D image space:
\begin{equation}
\text{MRE} = \frac{1}{NVJ} \sum_{i,j,v=1}^{N,J,V} \left\| \Pi_v(\xThreeDhat{i}{j}{}) - \xTwoD{i}{j}{v}{} \right\|_2,
\end{equation}
where $\Pi_v(\cdot)$ is the projection function using the intrinsic and extrinsic parameters of view $v$, and $\xTwoD{i}{j}{v}{}$ is the ground-truth 2D position in view $v$.

\noindent\textbf{Percentage of Correct Keypoints in 2D ($PCK_{2D}$)}  
$PCK_{2D}@\tau$ measures the fraction of 2D keypoints with reprojection error below a pixel threshold $\tau$:
\begin{equation}
PCK_{2D}@\tau = \frac{1}{NVJ} \sum_{i,j,v=1}^{N,J,V} \mathds{1}\biggr[ \left\| \Pi_v(\xThreeDhat{i}{j}{}) - \xTwoD{i}{j}{v}{} \right\|_2 < \tau \biggr]
\end{equation}
where $\mathds{1}[\cdot]$ is the indicator function. 
\\\\
\clearpage
\noindent\textbf{Percentage of Correct Keypoints in 3D ($PCK_{3D}$)}  
$PCK_{3D}@\delta$ evaluates the percentage of 3D joints with Euclidean error less than a threshold $\delta$ (in mm):
\begin{equation}
PCK_{3D}@\delta = \frac{1}{NJ} \sum_{i,j=1}^{N,J} \mathds{1}\biggr[ \left\| \xThreeDhat{i}{j}{} - \xThreeD{i}{j}{} \right\|_2 < \delta \biggr].
\end{equation}

\noindent\textbf{Mean Percentage of Correct Keypoints}
Additionally we use the expression mean PCK ($mPCK$) to denote the Mean Percentage of Correct Keypoints in either 2D or 3D. For 2D the threshold values are 5, 10, 20 and 30, and for 3D the values correspond to 5, 10, 25, 50, from which the mean is taken, similar to the evaluation protocol \cite{lin2015microsoftcococommonobjects}. The unit is either mm for $mPCK_{3D}$ and pixel for $mPCK_{2D}$.
\\\\
\noindent\textbf{Average Precision (AP) and Average Recall (AR)}  
To evaluate detection quality in 2D pose estimation, we also compute AP and AR scores across multiple keypoint thresholds, following the COCO evaluation protocol \cite{lin2015microsoftcococommonobjects}:
\begin{equation}
\text{AP} = \frac{1}{K} \sum_{k=1}^{K} \text{Precision}(k)
\end{equation}
where $K$ is the number of thresholds (from 0.5 to 0.95 in 0.05 increments), and Precision($k$) is  computed at each threshold using matching between predicted and ground truth keypoints based on Object Keypoint Similarity (OKS) reported in \cite{lin2015microsoftcococommonobjects}.
\\\\
\noindent\textbf{Mean Joint Error (MJE)}
We report mean joint error for the evaluation of 2D poses in the image space, giving the mean error for the hand poses given ground truth poses:
\begin{equation}
\text{MJE} = \frac{1}{NVJ} \sum_{i,j,v=1}^{N,J,V} \left\| \xTwoDhat{i}{j}{v}{} - \xTwoD{i}{j}{v}{} \right\|_2
\end{equation}

These metrics jointly provide a robust evaluation of the geometric accuracy of the estimated hand poses, both in image and 3D space for the full pipeline, as well as the detectability and localization precision of individual keypoints.

\clearpage
\section{Dataset}
\label{sec:dataset}

This section details the experimental datasets used to evaluate the proposed methods, encompassing both controlled non-invasive and anatomically realistic ex-vivo surgical scenarios.
\\\\
\noindent \textbf{SPP (Spinal Preparation; non-invasive mock patient)}
A resident orthopedic surgeon simulated the pre-incision phase of a midline lumbar approach on a volunteer mock patient. The sterile field was prepared with standardized surgical drapes (IVF Hartmann, Neuhausen am Rheinfall, Switzerland) and Ioban incise foil (3M, Maplewood, MN, USA). No cuts, resections, or instrumentation were performed. This scenario isolates lighting, draping, and staff-induced occlusions without tissue deformation from dissection, providing a controlled yet realistic background for pose-estimation benchmarking.
\\\\
\noindent \textbf{SPI (Spinal Instrumentation; ex-vivo)}
A cadaveric midline lumbar approach (fresh-frozen torso, Th1--coccyx; male, 76~years, BMI 21.7~kg/m$^2$) was performed by an attending surgeon with a resident orthopedic surgeon. The specimen (Science Care, Arizona, US) was stored at $-22^\circ$C and thawed for 72~h at room temperature at OR-X, Zurich, prior to use. The torso was placed on a commercial height-adjustable OR table (Hillrom, PST 500), draped as in SPP, and hydrated with periodic moist gauze coverage. The approach exposed L1--L5 with selective soft-tissue removal to access dorsal bony landmarks (mammillary processes, facet joints, laminae, spinous processes). Pedicle access followed routine practice using an awl and luer; Pennybacker and Cobb instruments were used as needed. Canal integrity was verified with a pedicle feeler before temporary placement of standardized polyaxial head cannulated pedicle screws ($6\times50$,mm) over a flexible K-wire. Instruments were provided by Ulrich AG (approach) and Medacta International (spinal instrumentation). This scenario introduces specular tools, transient occlusions, and view-dependent effects representative of clinical reality.

\subsection{Acquisition rigs and imaging parameters}
\label{sec:supp-acq}

\noindent \textbf{Camera settings}
For the SPP setup, 4 far-field and 8 near-field cameras (GoPro Hero~12 Black) were used.
For the SPI setup, 4 far-field and 7 near-field cameras (GoPro Hero~12 Black) were used.
All devices recorded at 4K resolution (9:16 aspect ratio) and 30~fps using the Linear lens mode.
The white balance was fixed at 4500~K for all cameras.
Far-field cameras operated with a shutter speed of 1/240~s and ISO~400.
For near-field cameras, ISO was set to 400 for SPP and 100 for SPI, with a fixed shutter speed of 1/960~s.
\\\\
\noindent \textbf{Camera intrinsics}
Intrinsic parameters were estimated using a professional checkerboard \cite{CalibIO2025}. Radial and tangential distortions were corrected, and undistorted images were used in all experiments.

\clearpage
\subsection{Scene selection}
We describe in the following section the selection of scenes used for analysis.
\\\\
\noindent \textbf{SPP1} Scene showing the surgeon performing the initial surgical approach. Standing on the right side of the patient, the surgeon uses a scalpel and surgical forceps to perform a superficial incision over the lumbar and thoracolumbar spine.
 \\\\
\noindent \textbf{SPP2} Scene showing the surgeon organizing instruments and ensuring safety. The surgeon collects and folds gauze on the instrument table, then places a safety cover over the electrocautery tip before positioning the instrument back near the patient.
\\\\
\noindent \textbf{SPP3} Scene showing the surgeon preparing for the surgical approach. The surgeon retrieves a scalpel and forceps from the instrument table and turns toward the patient’s right side to perform a superficial incision over the lumbar and thoracolumbar spine.
\\\\
\noindent \textbf{SPP4} Scene showing the surgeon positioning before the surgical approach. The surgeon moves along the patient’s right side, turns toward the operative field, and palpates the sterile area over the lumbar spine to confirm the surgical site.
\\\\
\noindent \textbf{SPP5} Scene showing the surgeon initiating the skin incision. Standing on the patient’s right side, the surgeon uses a scalpel and surgical forceps to perform the first superficial incision over the lumbar and thoracolumbar spine.
\\\\
\noindent \textbf{SPP6} Scene showing the surgeon completing the initial incision. Standing on the patient’s right side, the surgeon finalizes the first skin incision and performs additional small cuts at the incision site using a scalpel and surgical forceps over the lumbar and thoracolumbar spine.
\\\\
\noindent \textbf{SPP7} Scene showing the surgeon performing hemostasis. Standing on the patient’s right side, the surgeon uses surgical forceps to grasp bleeding vessels and applies a monopolar electrocautery instrument to the forceps tip to achieve coagulation.
\\\\
\noindent \textbf{SPI1} Scene showing the surgeon alternating between canal palpation and screw insertion preparation. The surgeon probes the pedicle canal for cortical breaches, retrieves the pedicle screwdriver with a mounted screw, verifies canal integrity once more, and positions the screwdriver at the predefined pedicle trajectory for screw placement.
\\\\
\noindent \textbf{SPI2} Scene showing the surgeon initiating pedicle screw insertion. The surgeon aligns the screwdriver with the pre-drilled pedicle trajectory while maintaining exposure with a Cobb elevator, briefly adjusts the elevator position, and then advances the pedicle screw into the vertebra.

\subsection{Ethics}
SPP involved a non-invasive mock patient; no protected health information is present. SPI was conducted under ethics approval (BASEC Nr.~2021-01196).

\subsection{Ground truth}
The ground truth was generated using a custom Python-based annotation tool with a graphical user interface. This software allowed synchronized viewing of multi-camera video streams, enabling annotators to select specific frames and manually label the left and right hands independently.

To compute the 3D ground truth, we jointly optimize the 3D joint locations obtained by triangulation of the manually annotated 2D hand keypoints across multiple camera views, minimizing both the reprojection errors and an additional regularization that penalizes deviations between the predicted joint-to-joint distances and their estimated nominal bone lengths. For each timestamp $t$, let $M_t$ denote the number of visible hands, and let $\mathcal{P}$ be the set of bone pairs $(j, k)$ defining the hand kinematic structure. For each hand $m$ and bone $(j, k)$, the term minimizes the squared difference between the predicted bone length $\|\xThreeDhat{t}{j}{h} - \xThreeDhat{t}{k}{h}\|_2$ and the estimated reference length $\hat{l}_{j,k}^{m}$. The total loss is averaged over all bones, hands, and timestamps belonging to sequences involving the same surgeon:
\begin{equation}
L_{\text{bone}} =
\frac{1}{\sum_{t=1}^{T} M_t\,|\mathcal{P}|}
\sum_{t=1}^{T} \sum_{m=1}^{M_t} \sum_{(j,k)\in\mathcal{B}}
\Big(\big\|\xThreeDhat{t}{j}{m}-\xThreeDhat{t}{k}{m}\big\|_2 - \hat{l}_{j,k}^{m}\Big)^2.
\end{equation}
This term enforces consistent bone lengths across all timestamps and sequences recorded from the same surgeon, improving anatomical plausibility.
Optimization is performed using the L-BFGS algorithm with strong Wolfe line search.
The hyperparameters used were respectively set to $\lambda_{\text{reproj}} = 1$ and $\lambda_{\text{bone}} = 100$.
\\\\
\noindent \textbf{Inter-annotator variability} To quantify the accuracy of our ground truth poses, we evaluated inter-annotator variability across 2D and 3D hand pose annotations. Four annotators independently labeled a total of 558 hands corresponding to six randomly selected timestamps and seen by 8 cameras among five (SPP1 -- SPP5) sequences. For each annotated frame, pairwise Euclidean distances were computed between all annotator pairs for each corresponding joint, yielding both 2D pixel distances (after re-projection to camera space) and 3D distances (in millimeters) in the global coordinate system. These distances were then averaged across all frames and cameras to obtain an overall measure of variability per joint.

\begin{figure}[h!]
    \centering
        \begin{subfigure}[t]{0.32\textwidth}
            \centering
            \includegraphics[width=\linewidth]{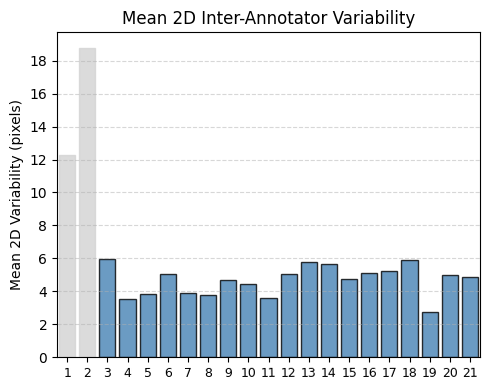}
            \caption{2D inter-annotator variability per joint (in pixels).}
            \label{fig:inter_var_2d}
        \end{subfigure}
        \hfill
        \begin{subfigure}[t]{0.32\textwidth}
            \centering
            \includegraphics[width=\linewidth]{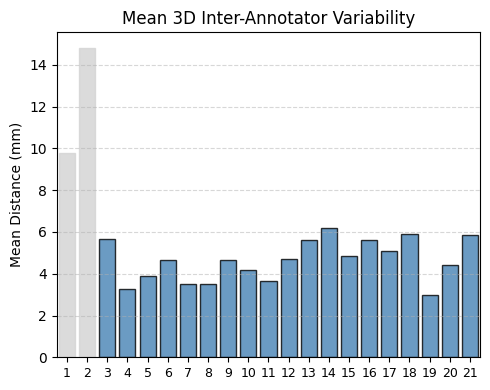}
            \caption{3D inter-annotator variability per joint (in mm).}
            \label{fig:inter_var_3d}
        \end{subfigure}
        \hfill
    \begin{subfigure}[t]{0.32\textwidth}
        \centering
        \includegraphics[height=0.8\linewidth, keepaspectratio]{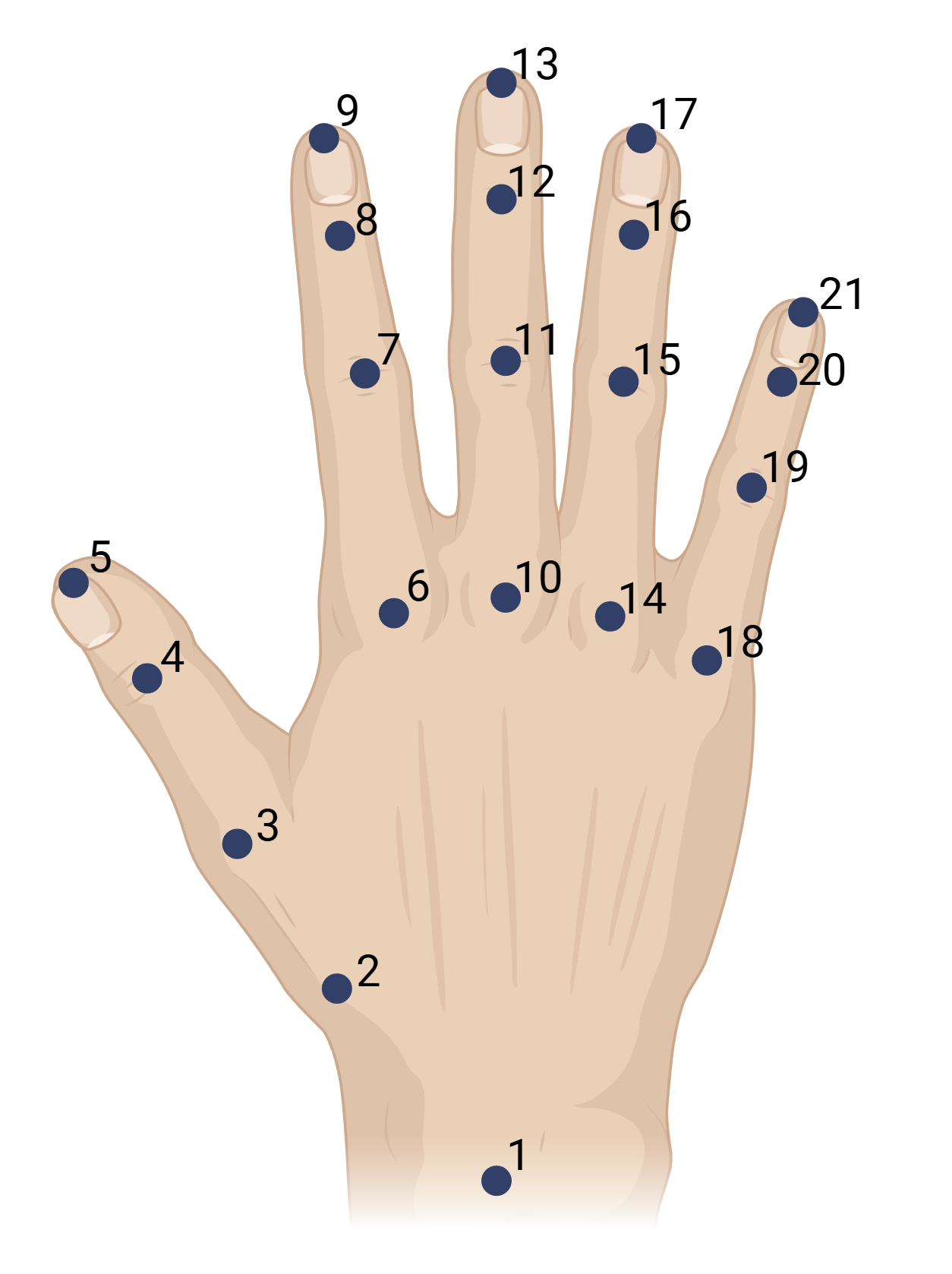}
        \caption{Hand joint model with 21 annotated keypoints. Created in BioRender.}
        \label{fig:hand_skeleton}
    \end{subfigure}

    \caption{
        \textbf{Inter-annotator variability across 2D and 3D annotations.}
        (a) 2D pixel-space variability aggregated across all eight cameras.
        (b) Mean 3D pairwise distances between all annotator pairs, computed per joint and averaged across all sequences and frames.
        (c) Illustration of the 21-joint hand model used for labeling and distance computation.
    }
    \label{fig:inter_annotator_variability}
\end{figure}

The results are shown in Fig.~\ref{fig:inter_annotator_variability}. Except for joints~1 and~2 (wrist and base of the palm), the inter-annotator variability remains below 6~pixels in~2D and below 6~mm in~3D, indicating high annotation consistency and labeling reliability. Joints~1 and~2, highlighted in grey in the histograms, were discarded from the ground truth due to their higher variability, which reflects inconsistent labeling unsuitable for quantitative evaluation.

Both the annotation software and the dataset will be released as open-source resources.

\section{Additional results}
\label{sec:additional_results}

This section presents additional quantitative and qualitative results that further demonstrate the performance and limitations of the proposed method.

Table \ref{tab:full_pipeline_results} presents per-sequence quantitative results of the full pipeline. Each column of the table corresponds to a specific sequence among SPP or SPI and complexity level (\myL{0} – \myL{2}). The reported metrics include the mean per-joint position error (MPJPE, in millimeters), mean reprojection error (MRE, in pixels), and the mean percentage of correct keypoints in 2D and 3D ($mPCK_{2D}$, $mPCK_{3D}$). Lower values of MPJPE and MRE indicate better spatial and projection accuracy, while higher $mPCK$ values reflect improved pose estimation consistency.

\begin{figure}[t]
    \centering
    \includegraphics[width=\linewidth]{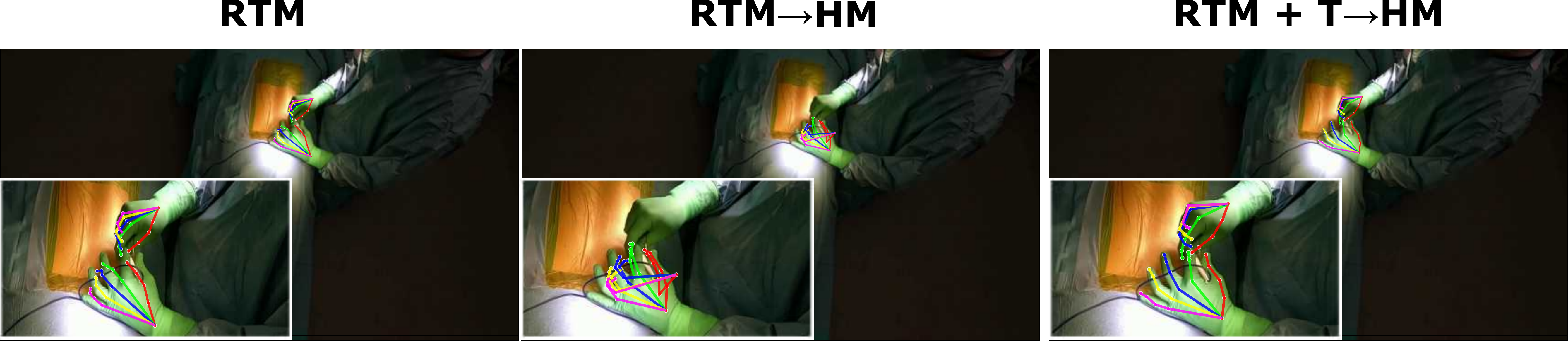}
    \caption{Qualitative comparison of 3D hand poses estimated using baseline methods and our approach (shown reprojected onto input images). From left to right: RTM Whole-Body, RTM~$\rightarrow$~HM, and our method combining intermittent RTM, hand tracking, and HM. The first detects both hands but with coarse joints, the second misses one hand and mislocalizes joints, while ours accurately detects and localizes both hands.}
    \label{fig:abl_fig}
\end{figure}

To highlight the improvements achieved by our full pipeline, Fig.~\ref{fig:abl_fig} presents a qualitative comparison of 3D hand pose estimations from baseline methods and our approach.

\begin{figure}[t]
    \centering
    \includegraphics[width=\linewidth]{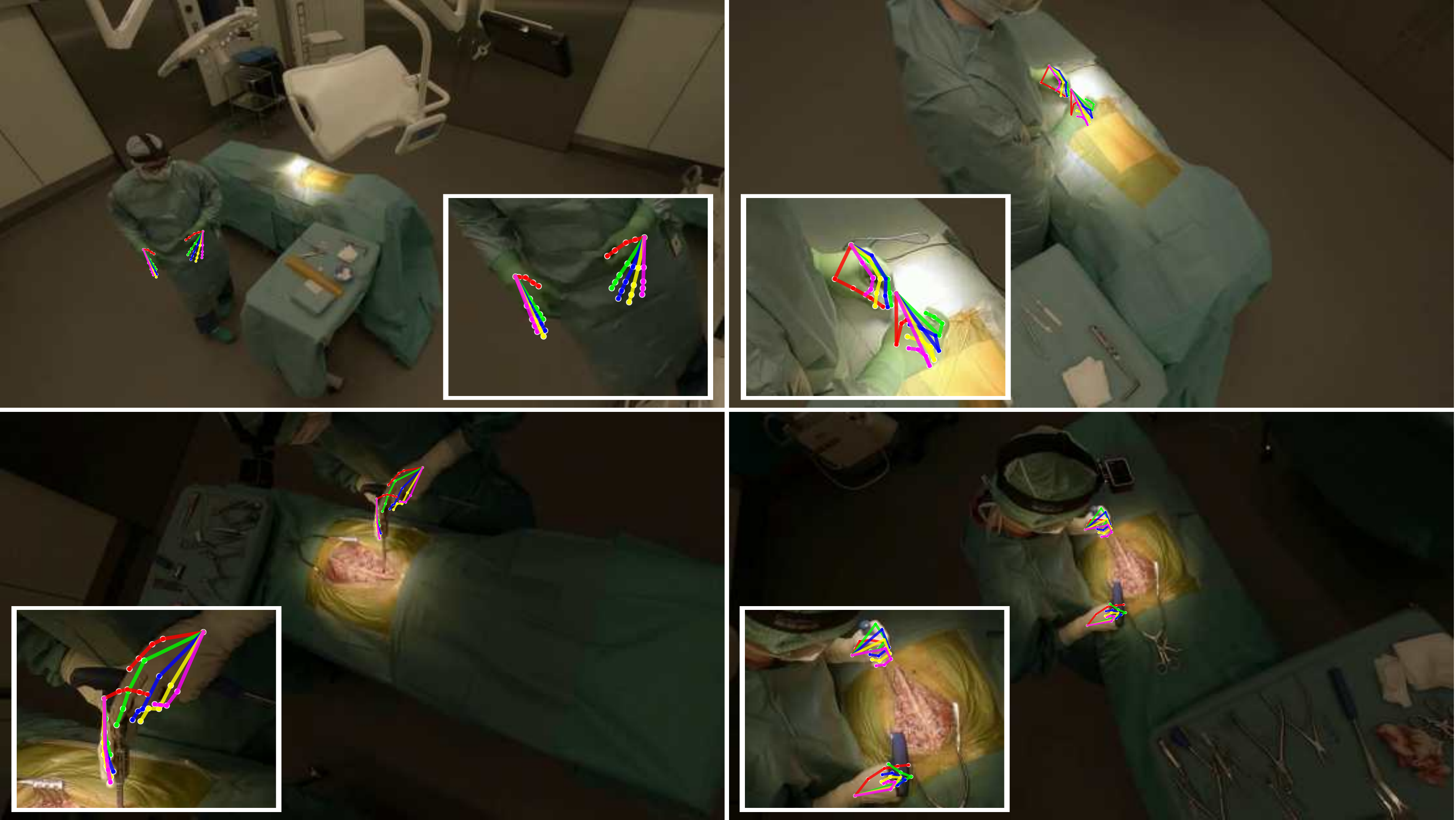}   
    \caption{The main causes of failure include severe occlusions from surgical instruments or other objects, limited camera coverage (fewer than three active views), and strongly self-occluding hand poses such as closed fists.}
    \label{fig:limitation}
\end{figure}

To illustrate some limitations of our approach, Fig.~\ref{fig:limitation} presents four qualitative examples of failure cases observed under challenging conditions (see caption for details).

\begin{table}[t]
\setlength{\tabcolsep}{3pt}
\centering
\rowcolors{2}{gray!10}{white} 
\resizebox{\linewidth}{!}{
    \begin{tabular}{lccccccc|cc|c}
    \toprule
    \textbf{Data Collection} & \mycha 1 & \mycha 2 & \mycha 3 & \mycha 4 & \mycha 5 & \mycha 6 & \mycha 7 & \mychb 1 & \mychb 2 & \textbf{Mean}\\
    \textbf{Complexity} & \myL{0} & \myL{1} & \myL{1} & \myL{1} & \myL{0} & \myL{0} & \myL{0} & \myL{2} & \myL{2} & \\
    \midrule
    \midrule
    \textbf{MPJPE (mm)} $\downarrow$ & 8.3 & 9.1 & 10.2 & 8.1 & 8.8 & 8.5 & 7.0 & 7.6 & 8.8 & 8.5\\
    \textbf{MRE (px)} $\downarrow$  & 6.0 & 8.2 & 9.0 & 8.7 & 7.3 & 7.2 & 6.9 & 9.8 & 10.3 & 8.2\\
    $mPCK_{2D}$ $\uparrow$  & 83.6 & 74.6 & 74.9 & 76.0 & 77.9 & 78.8 & 79.9 & 72.9 & 70.3 & 76.6\\
    $mPCK_{3D}$ $\uparrow$  & 76.1 & 70.7 & 71.8 & 77.2 & 71.7 & 74.0 & 79.8 & 77.5 & 73.4 & 74.7\\
    \bottomrule
    \end{tabular}
}
\caption{Full pipeline results on the surgical dataset showing MPJPE (mm), reprojection errrors (px), and mean PCK (\%) of differing scenarions.}
\label{tab:full_pipeline_results}
\end{table}

\section{Implementation details}
\label{sec:impl_details}

All components of the proposed pipeline are built from publicly available pretrained models and run without any fine-tuning. Below, we summarize the key implementation details and hyperparameters for reproducibility.

\noindent \textbf{2D detection and initial estimates.} Person detection is performed using YOLOv11x~\cite{jocher_2025_15616841}. As mentioned in the main paper, we assume a single-person scenario; therefore, for each frame, only the highest-confidence bounding box is retained for further processing. The RTMPose-Wholebody pose estimation model is then applied to these detections.
\\\\
\noindent \textbf{Wrist confidence score}
For wrist keypoints, the confidence scores are set to $1.0$, as empirical observations show that wrist detections are highly stable across frames. In the 3D optimization (Eq.~1 in the main paper), these confidence values weight the reprojection loss term; assigning a fixed high confidence effectively anchors the hand to the wrist position. This prevents temporal smoothness regularization from dominating during fast hand motions, ensuring accurate and stable wrist reconstruction.
\\\\
\noindent \textbf{3D pose estimation}
We employ the L-BFGS-B~\cite{lbfgsb} optimizer (PyTorch v2.5.1) with a learning rate of 1.0, a maximum of 100 iterations, and a convergence tolerance of $1 \times 10^{-5}$.
Optimization is performed on 50-frame windows with a 25-frame overlap to ensure temporal consistency.
In our ablation study, the weighting factors are set to $\lambda_{\text{repr}} = 1.0$, $\lambda_{\text{smooth}} = 20.0$, $\lambda_{\text{shape}} = 50.0$, and $\lambda_{\text{biomech}} = 5.0$.
For the final configuration, we exclude the biomechanical term ($\lambda_{\text{biomech}} = 0$), as it was found to significantly decrease 3D pose accuracy.

\end{document}